\documentclass[conference]{IEEEtran}
\IEEEoverridecommandlockouts
\usepackage{cite}
\usepackage{amsmath,amssymb,amsfonts}
\usepackage{algorithmic}
\usepackage{graphicx}
\usepackage{textcomp}
\usepackage{xcolor}
\usepackage{subcaption}
\def\BibTeX{{\rm B\kern-.05em{\sc i\kern-.025em b}\kern-.08em
    T\kern-.1667em\lower.7ex\hbox{E}\kern-.125emX}}
\begin{document}

\title{High-Throughput and Accurate 3D Scanning of Cattle Using Time-of-Flight Sensors and Deep Learning\\

}


\author{
    \IEEEauthorblockN{Gbenga Omotara\IEEEauthorrefmark{1}, Seyed Mohamad Ali Tousi\IEEEauthorrefmark{1}, Jared Decker\IEEEauthorrefmark{2}, Derek Brake\IEEEauthorrefmark{2}, Guilherme N. DeSouza\IEEEauthorrefmark{1}}
    \IEEEauthorblockA{\IEEEauthorrefmark{1}ViGIR Lab, EECS Dpt., University of Missouri, Columbia, US\\ }
    \IEEEauthorblockA{\IEEEauthorrefmark{2}Division of Animal Sciences, University of Missouri, Columbia, US\\ Email: {goowfd, stousi}@mail.missouri.edu,{ deckerje, braked, desouzag}@missouri.edu}
}

\maketitle

\begin{abstract}
We introduce a high throughput 3D scanning solution specifically designed to precisely measure cattle phenotypes. This scanner leverages an array of depth sensors, i.e. time-of-flight (Tof) sensors, each governed by dedicated embedded devices. The system excels at generating high-fidelity 3D point clouds, thus facilitating an accurate mesh that faithfully reconstructs the cattle geometry on the fly.
In order to evaluate the performance of our system, we have implemented a two-fold validation process. Initially, we test the scanner's competency in determining volume and surface area measurements within a controlled environment featuring known objects. Secondly, we explore the impact and necessity of multi-device synchronization when operating a series of time-of-flight sensors. Based on the experimental results, the proposed system is capable of producing high-quality meshes of untamed cattle for livestock studies. 


\end{abstract}


\section{Introduction}
On November 15, 2022, the world's population surpassed the milestone of 8 billion people \cite{owidpopulationgrowth}, thereby accentuating the significance of food production on a global scale. Reliable quantification of livestock and plant phenotypes has emerged as a vital factor in maximizing efficiency and sustainability in these fields.

Cattle farming forms an integral part of global agriculture, serving as a substantial contributor to food resources and economic stability for numerous communities. In the wake of technological advancements, we see increasing adoption of innovative practices aimed at enhancing productivity and sustainability in cattle farming. One such technological breakthrough is 3D cattle scanning, which presents a comprehensive three-dimensional rendition of bovine bodies.

Utilizing 3D scanning to capture the volume and surface area of cattle enables farmers and researchers to extract valuable insights about the animals' health, growth trends, and overall welfare. Such precise measurements aid in the early detection of potential health issues or discomfort zones, paving the way for timely interventions and treatments. Additionally, it facilitates accurate body condition scoring, which supports optimized feed and nutrition management strategies.

Aside from its practical applications, 3D cattle scanning is an important research tool within the field of animal science. Researchers can gain insight into cattle genetics, breeding patterns, behavior, and physiology by examining data obtained from 3D scans. Cattle farmers will benefit from these findings in terms of establishing novel breeding programs, improving animal welfare, and bolstering sustainability and profitability.

Here, we introduce a 3D cattle scanning system that utilizes RGBD cameras to produce precise and reliable 3D models of cattle. The system is engineered to automate the process of calculating the animals' volume and surface area. This paper is structured as follows: the second section pays homage to valuable existing work in the domain of 3D animal scanning, 3D point cloud registration, 3D surface reconstruction, and calculation, as well as conventional cattle phenotyping methodologies. The third section provides a detailed overview of the proposed scanning system, explaining the methods employed within its pipeline. The fourth section exhibits the experiments conducted and the results derived from the system's implementation and evaluation. The final section delves into a comprehensive discussion of the results and the system's performance in diverse working environments, along with a conclusion on the proposed method.

\begin{figure*}[t]
    \centering
    \includegraphics[width = 180mm]{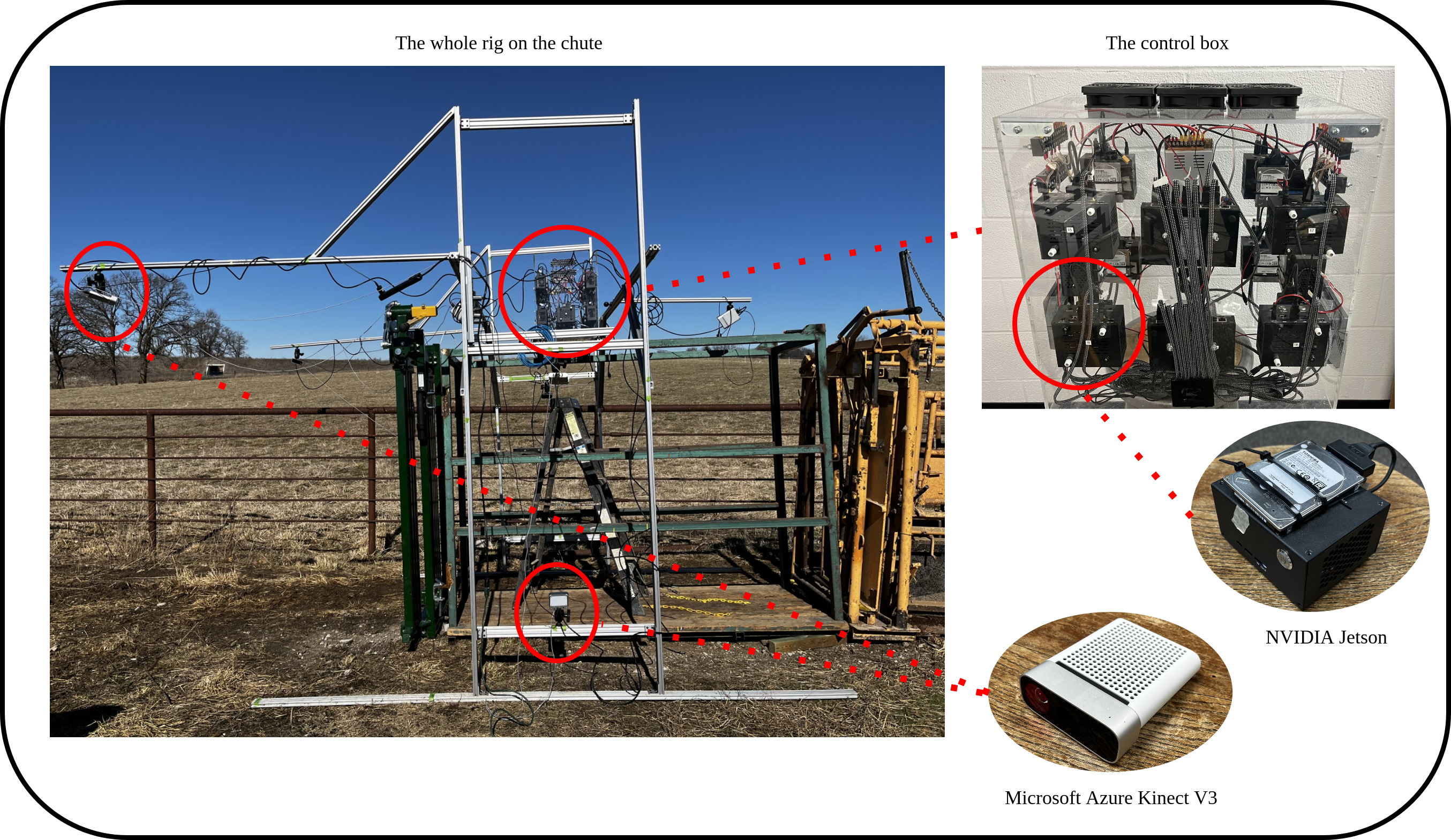}
    \caption{The proposed platform for 3D cattle scanner system.}
    \label{fig:platform}
\end{figure*}

\section{Background}
\subsection{Scanners for Phenotyping large Animals}
The field of using computer vision for the measurement of livestock physical traits is rapidly expanding, with numerous studies contributing significant advancements.

For instance, a study conducted in 2013 utilized three Kinect V1 sensors to assess the quality of Japanese Black cattle, estimating their weight, size, shape, and posture during growth \cite{KawasueThreeDimensionalSM}. In another study \cite{article}, researchers developed a handheld 3D scanner combined with a 3D Thermo-sensing device for monitoring Japanese Black cattle by aligning both depth and thermal measurements.

Further advancements were made in \cite{Ruchay_2019}, where  the capability of a sensor system was demonstrated under three distinct scenarios: 1) Several cameras mounted on a moving platform revolving around the subject, with known camera positions, 2) a handheld scanning device, and 3) multiple Kinect cameras secured on a platform through which the animal passes.

Another approach was presented in \cite{7791955} where authors utilized eight RGB cameras mounted on four pole stands, with control managed by a Raspberry Pi device. They used the Shape from Silhouette method to carve out the cattle's geometry, measuring cattle volume by counting the number of generated points. The authors highlighted the need for sensor synchronization which usually goes unmentioned. A different approach was taken in \cite{WANG2018291}, where two Xtion Pro cameras were used to capture 3D data from pigs, enabling the estimation of pig body measurements such as size. The study in \cite{LI2022106987} introduced a multi-view setup with 5 hardware synchronized Azure Kinect cameras. These cameras are arrayed across a gantry structure to generate 3D point clouds of cattle. In this work, the authors conduct an analysis of the impact of light intensity when performing these scans, a very relevant observation since many of these systems need to be operated outdoors where the lighting conditions vary.

While these studies represent significant advancements in the field, there remain several areas for improvement and exploration, particularly in terms of scalability, and the issue of having untamed cattle, which our study aims to address. 

\section{Proposed Scanning System}

Our work contributes to the growing body of research presented in the previous section, by presenting a novel sensor rig comprised of eight Time-of-Flight Sensors arranged on a frame encircling a chute suitable for containing untamed cattle. Each Tof sensor, controlled by an embedded device specifically, is hardware-synchronized. We demonstrate that our sensor system can produce high-fidelity scans in real time. This efficiency is largely due to the utilization of a sophisticated image segmentation model, which automates the segmentation of cattle within both depth and RGB images.

In this section, we will describe our proposed platform to build the 3D cattle scanner system and its hardware and software implementation details. At the end of the section, we will also present our system's advantages over other scanning systems. 

\subsection{Platform}

Figure \ref{fig:platform} shows the platform being used to build our 3D cattle scanner. The frame is completely isolated from the chute to prevent any disturbances due to the animals' movements. A control box is responsible for controlling the cameras and taking images of the cattle which will be described in detail in what follows. 

\begin{figure*}
    \centering
    \includegraphics[width = 180mm]{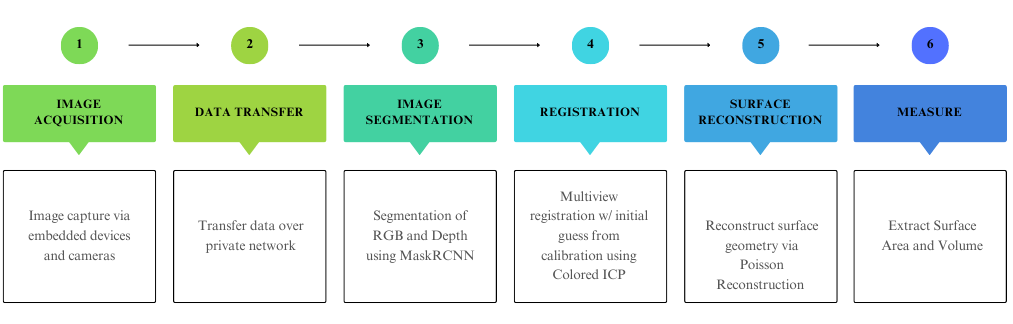}
    \caption{The block diagram of the proposed pipeline.}
    \label{fig:block}
\end{figure*}

\subsubsection{Hardware}

The scanner hardware implementation consists of three main parts; the camera setup, the control box, and the user laptop. The camera setup forms a space between the cameras to cover the whole surface of the animal. The current setup in the field is as follows:
\begin{enumerate}
    \item 2 cameras on both sides to cover the head.
    \item 2 cameras on top of the animal.
    \item 4 cameras on the sides to cover the animal's body.
\end{enumerate}

The automatic point cloud registration, described in the software section, allows a flexible camera setup that is agnostic to any kind of arrangement. For example, the setup varies between experimental use cases involving known objects and practical applications, like real-world cattle scanning.

The control box represents the next component of the scanner system. As depicted in figure \ref{fig:platform}, the control box houses the embedded devices, a pair of networking hubs, and the power distribution unit. Each embedded device is paired with an individual Tof sensor and manages the image acquisition of its respective camera. All elements within the control block are linked via a private network to the user's laptop, enabling control over the scanning procedure.

\begin{figure*}
    \centering
    \includegraphics[width = 180mm]{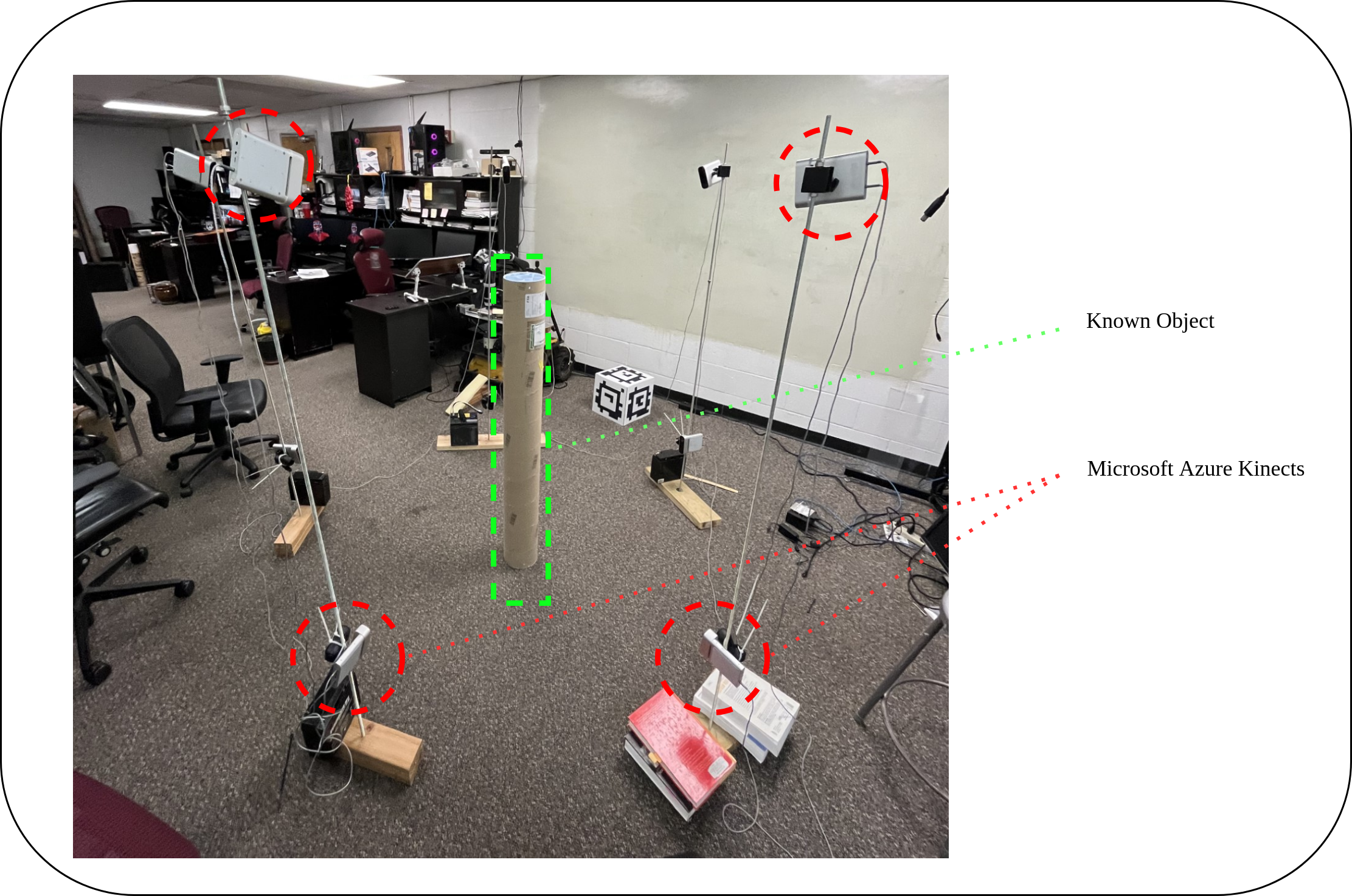}
    \caption{The camera setup for experimenting with the known objects and laser interference.}
    \label{fig:rig2}
\end{figure*}

\subsubsection{Software and Pipeline}

As depicted in figure \ref{fig:block}, the block diagram outlines the proposed software pipeline of the scanner. The scanning process starts with the acquisition of RGBD (color and depth) images of the animal, captured by the entire camera assembly. A user-friendly software has been designed, providing operators the ability to customize settings for the image acquisition process, such as automated or manual assignment of cattle IDs. This software, henceforth known as the Client program, establishes connections with all the embedded devices. Each embedded device hosts a Server program that processes requests from the Client and orchestrates the data acquisition process at a software level. Socket connections enable communication between the Client and the Server programs.

Following the initial image acquisition phase, all RGB and depth images are saved locally on the embedded devices. These images are then transmitted to the operator's laptop via a private network for subsequent processing.

The next step in the pipeline involves processing the RGB and depth images to segment the animal's body in preparation for the registration phase (stitching all the 3D point clouds. To accomplish this, a state-of-the-art deep learning segmentation model known as Mask R-CNN is utilized \cite{8237584}. The Mask R-CNN (Region-based Convolutional Neural Network) is renowned for its efficacy in tasks such as instance segmentation, object detection, and object recognition. The unified architecture of Mask R-CNN allows for the accurate identification and localization of multiple objects within intricate scenes, making it an ideal solution for tasks demanding a detailed instance-level understanding of visual data.

Originally trained by the Facebook AI Research (FAIR) team on the COCO (Common Objects in Context) data set \cite{10.1007/978-3-319-10602-1_48}, the Mask R-CNN model has been fine-tuned on a cattle-specific data set curated by the authors. In our work, we fine-tuned the model on both color and depth images which was inspired by the work in \cite{gumeli2023objectmatch}. This fine-tuning enhances the network's performance for the specialized task of cattle segmentation.

Upon completing the segmentation stage, the 2D coordinates of the segmented object are back-projected into 3D space to generate a segmented point cloud for each sensor. We then stitch these multiple 3D point clouds to create a single 3D model of the cattle.

The pipeline concludes with a 3D surface reconstruction performed on the registered point cloud. This enables the system to measure the surface area and volume of the cattle. The Poisson Surface Reconstruction algorithm is used for this purpose, which is designed to reconstruct a smooth, watertight surface from a disorganized collection of 3D points \cite{kazhdan2006poisson}.

\subsection{Advantages over Other Systems}

Our scanning system offers three distinct advantages over comparable systems:
\begin{enumerate}
    \item \textbf{Mobility}: The automatic point cloud registration provides operators with the flexibility to disassemble and reassemble the camera set in any configuration they desire. This versatility facilitates the system's transport, making it both possible and practical.

    \item \textbf{Scanning Speed}: By employing multiple advanced algorithms to process each cattle's data, such as a deep learning-based segmentation model and an autonomous image point cloud registration method, the system accelerates data acquisition and processing. This allows operators to scan numerous animals within a relatively short time span.

    \item \textbf{Ability to Scan Untamed Animals}:  Scanning untamed animals, which are considerably more challenging to restrain and scan, necessitates a robust and narrow chute to keep the animal still during the data capture. Such narrow chutes can complicate the point cloud segmentation of the animal. However, the fine-tuned deep learning-based segmentation model employed in this system is capable of handling all chute types, effectively and accurately segmenting the cattle from the chute body. 
\end{enumerate}

\section{Experimental Results and Discussions}

In this work, we used the Microsoft Azure Kinect as our depth sensor and the NVIDIA Jetson Nano as the controlling embedded system. All the cameras are subject to hardware synchronization. A series of 3.5mm audio cables are used to connect all cameras in a daisy chain fashion, leveraging the external synchronization features available with the depth sensors we use. The registration steps also were performed using Open3D library \cite{Zhou2018}. 
We designed our experiments in order to assess the viability of our sensor rig for the given task of scanning animals. Multiple experiments have been conducted to validate the proposed scanning system which will be presented in this section. 

\subsection{Implementation Considerations: Sensor Synchronization}

\begin{figure*}
     \centering
     \begin{subfigure}[b]{0.22\textwidth}
         \centering
         \includegraphics[width=\textwidth]{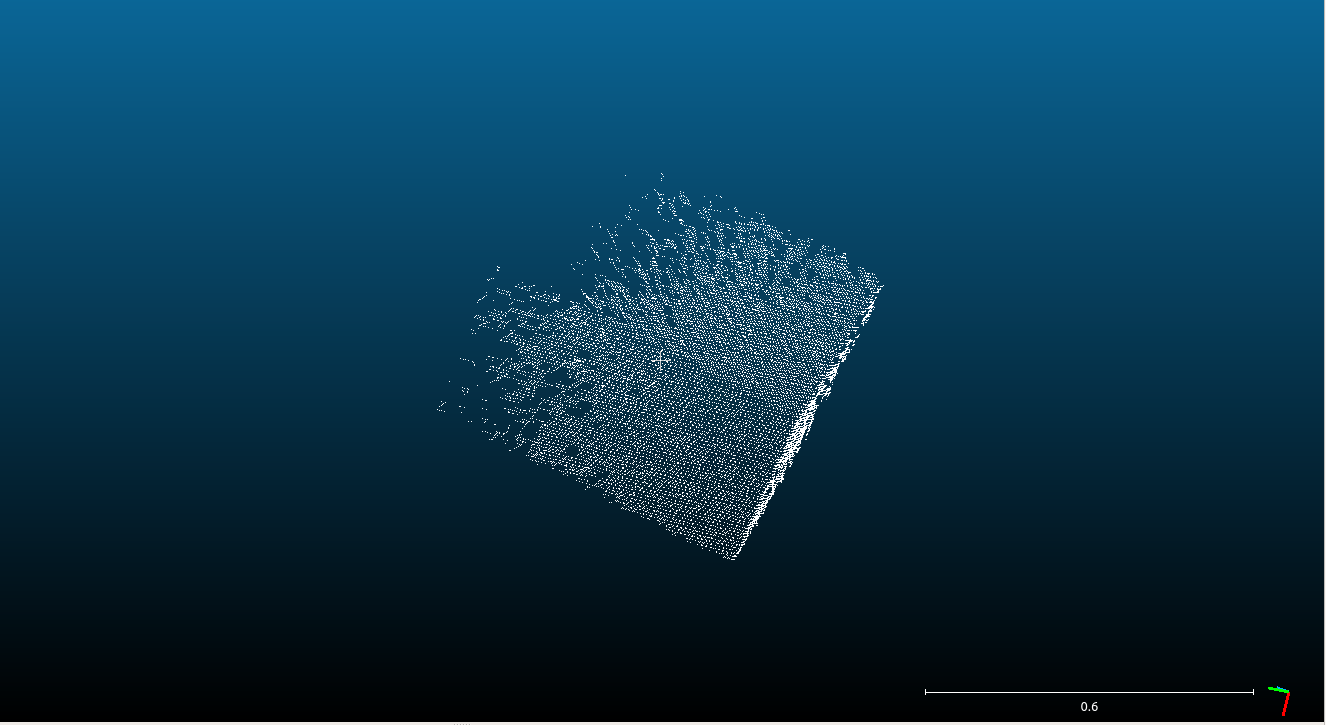}
         \caption{Medium box, 0s delay}
         \label{fig:segres1}
     \end{subfigure}
     \hfill
     \begin{subfigure}[b]{0.22\textwidth}
         \centering
         \includegraphics[width=\textwidth]{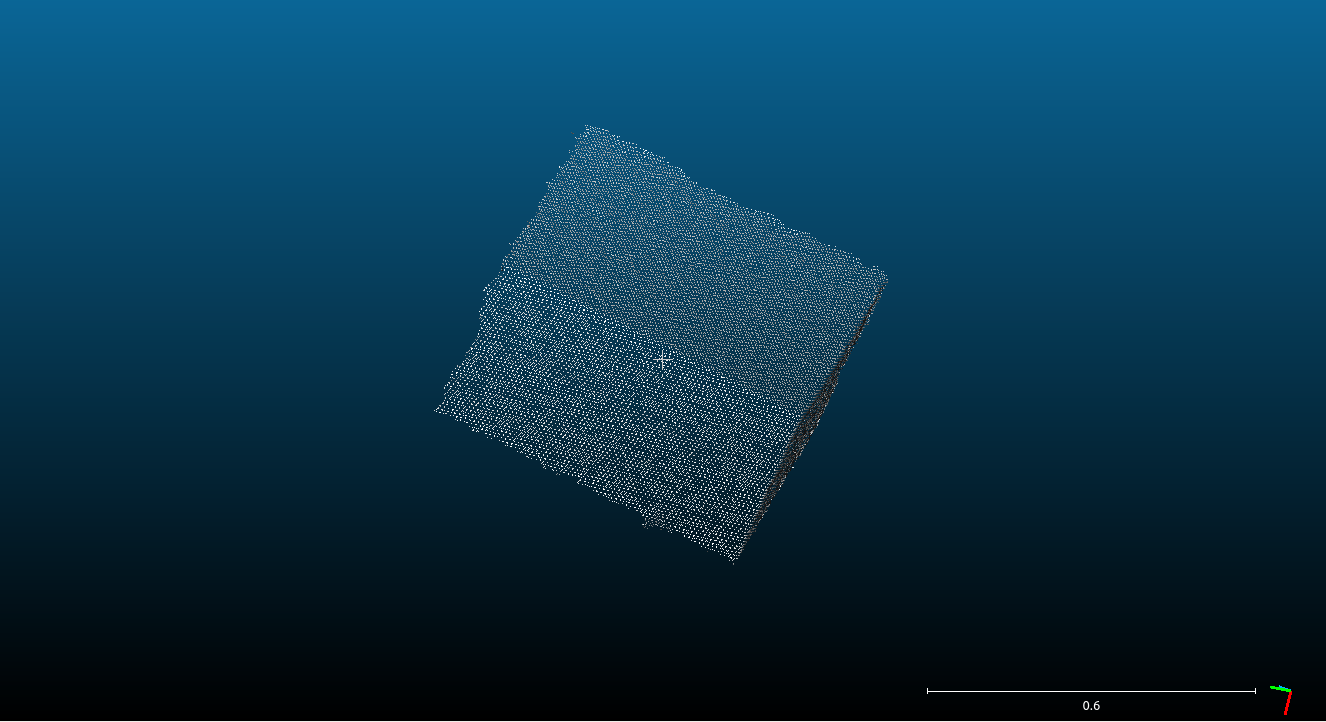}
         \caption{Medium box, 160 us delay}
         \label{fig:segres2}
     \end{subfigure}
     \hfill
     \begin{subfigure}[b]{0.22\textwidth}
         \centering
         \includegraphics[width=\textwidth]{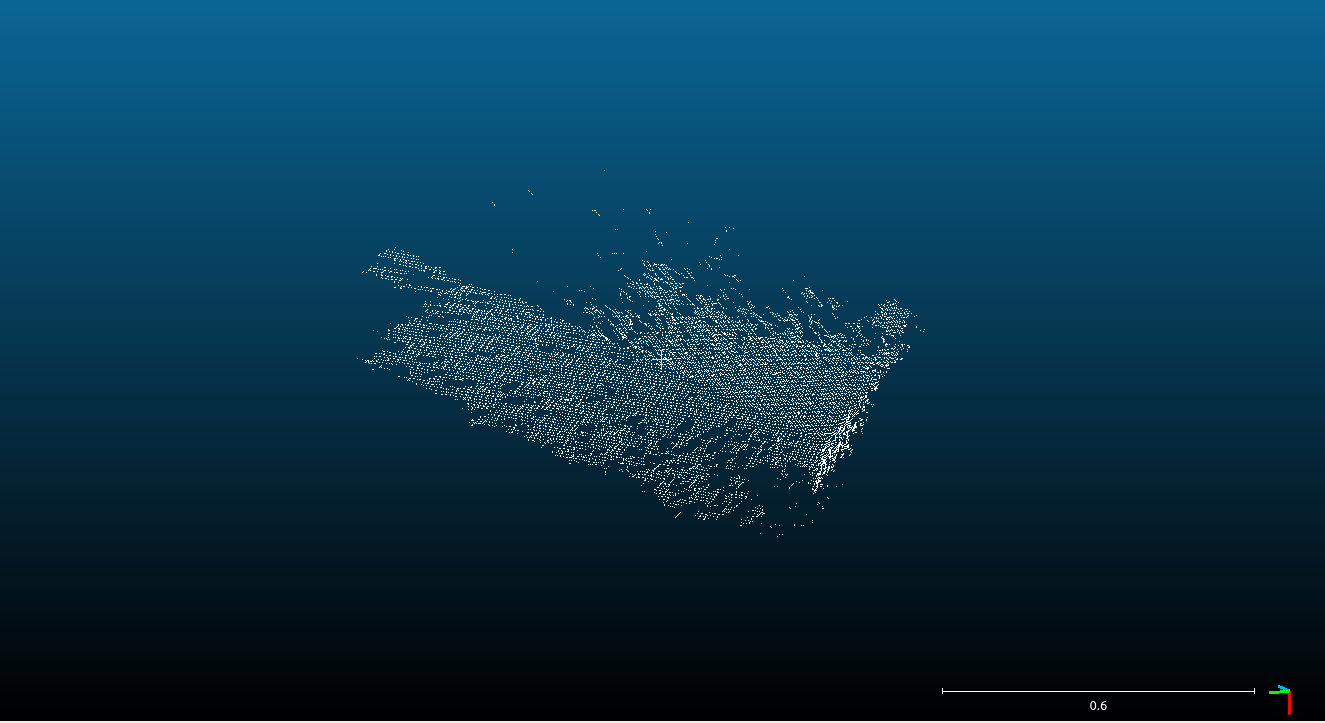}
         \caption{Large box, 0s delay}
         \label{fig:segres1}
     \end{subfigure}
     \hfill
     \begin{subfigure}[b]{0.22\textwidth}
         \centering
         \includegraphics[width=\textwidth]{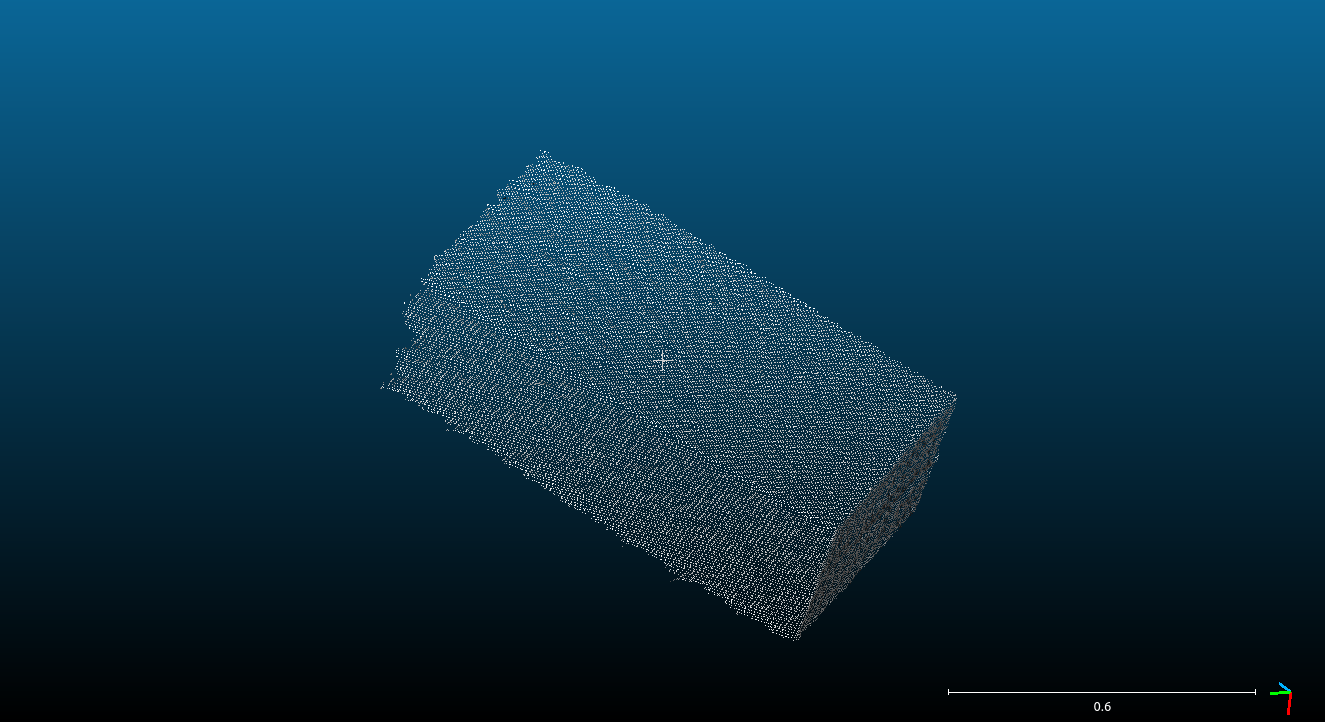}
         \caption{Large box, 160 us delay}
         \label{fig:segres1}
     \end{subfigure}
     \hfill
     \caption{A qualitative comparison between the scanner outputs to demonstrate the effect of using and not using the proper synchronization delay.}
     \label{fig:delayres}
\end{figure*}

The Tof sensor performs laser scans by casting modulated illumination in the near-IR spectrum onto the scene of interest. Given that we have a multi-sensor rig, these infrared signals can interfere with each other if they are simultaneously emitted. So substantial attention is needed on the capture timing and synchronization of the cameras. To demonstrate the effects of having and not having a reasonable synchronization in the data collection process, an experiment has been executed. In order to perform these tests, 10 Tof sensors have been assembled in a pairwise fashion on a rod as seen in Figure \ref{fig:rig2}. The tests were performed using three boxes with known dimensions. These known objects have been captured from all devices (N=10 views) in two operating modes, with and without hardware synchronization. The qualitative results are shown in Figure \ref{fig:delayres}. It is clearly observable that the effect of proper synchronization is severe and not having it can cause the system to lose more than 80\% of the object's points.

\subsection{Experiments on Known Objects}

\begin{figure}
     \centering
     \begin{subfigure}[b]{0.4\textwidth}
         \centering
         \includegraphics[width=\textwidth]{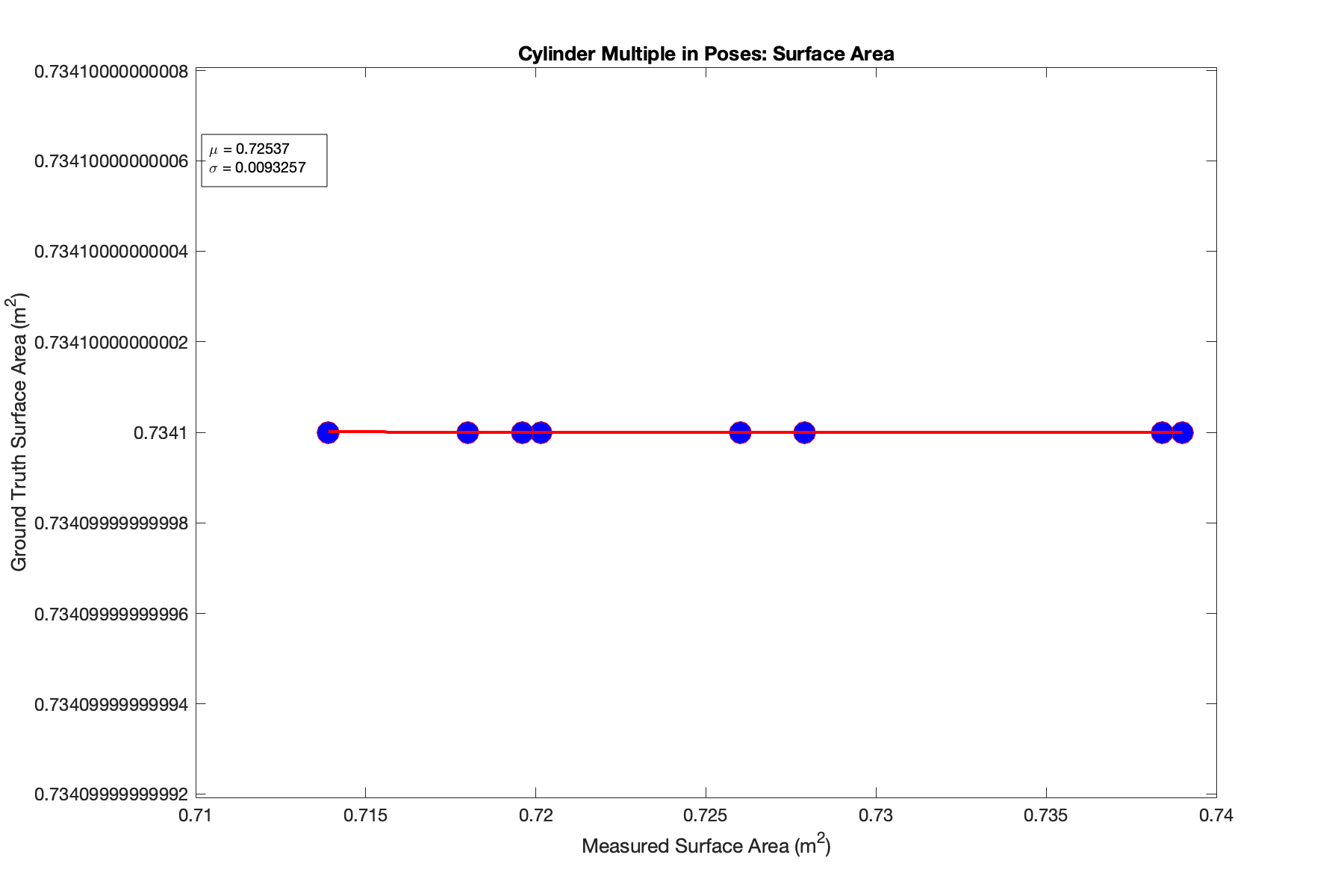}
         \caption{Surface area calculation results}
         \label{fig:cylcurres}
     \end{subfigure}
     \hfill
     \begin{subfigure}[b]{0.4\textwidth}
         \centering
         \includegraphics[width=\textwidth]{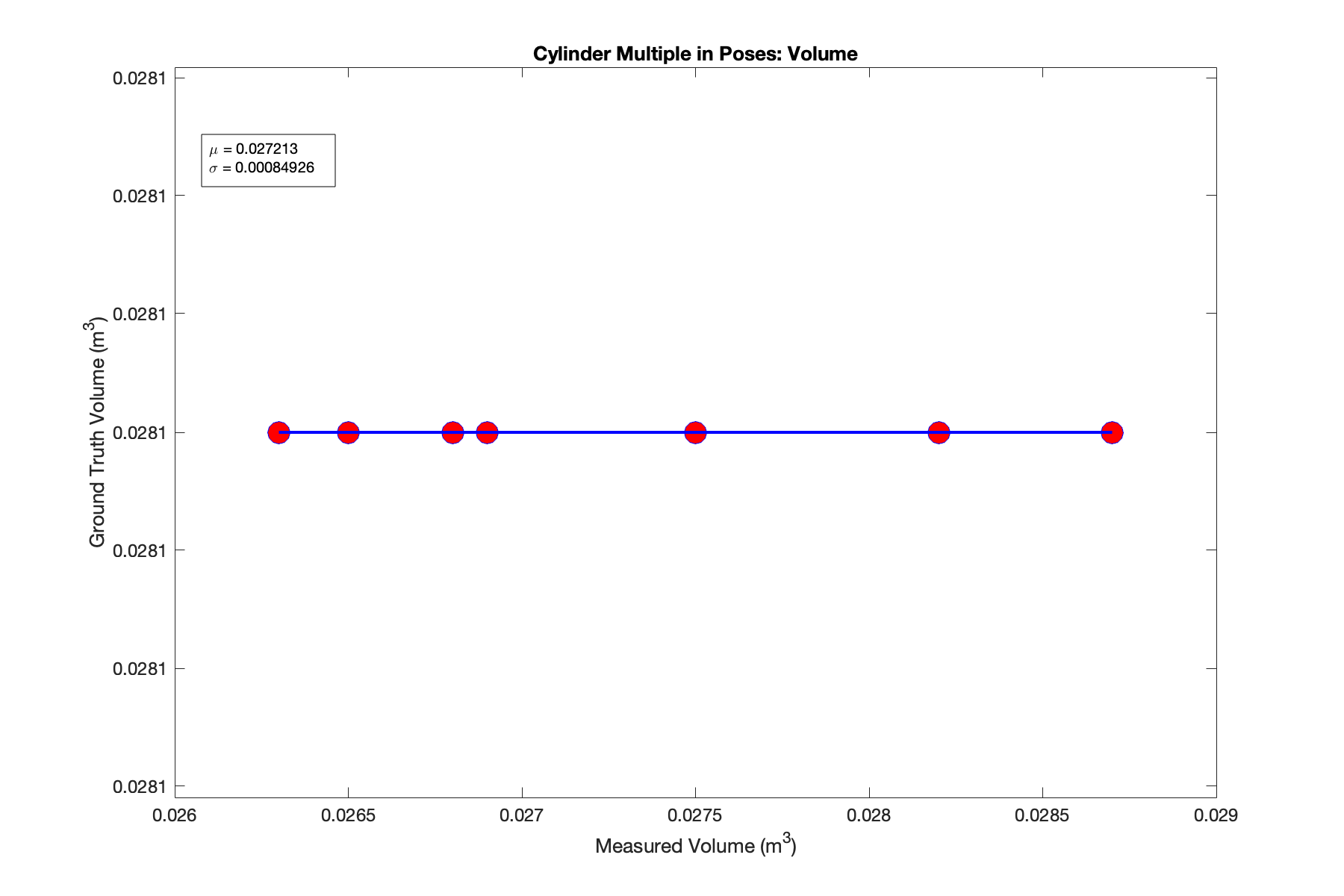}
         \caption{Volume calculation results}
         \label{fig:cylvolres}
     \end{subfigure}
     \hfill
     \caption{The statistical results of scanning one known object (a cylinder) in different orientations.}
     \label{fig:cylres}
\end{figure}

To validate the proposed scanning system's accuracy in determining the volume and surface area of the target object, several experiments were conducted. Four different objects, including three boxes and one cylinder, were used for testing. Figure \ref{fig:rig2} shows the setup used to test the scanner on the cylinder object. The cylinder was placed in various orientations, and the results, shown in Figure \ref{fig:cylres}, indicate that the scanner accurately calculated the surface area and volume of the cylinder in all orientations.

\begin{figure}
     \centering
     \begin{subfigure}[b]{0.4\textwidth}
         \centering
         \includegraphics[width=\textwidth]{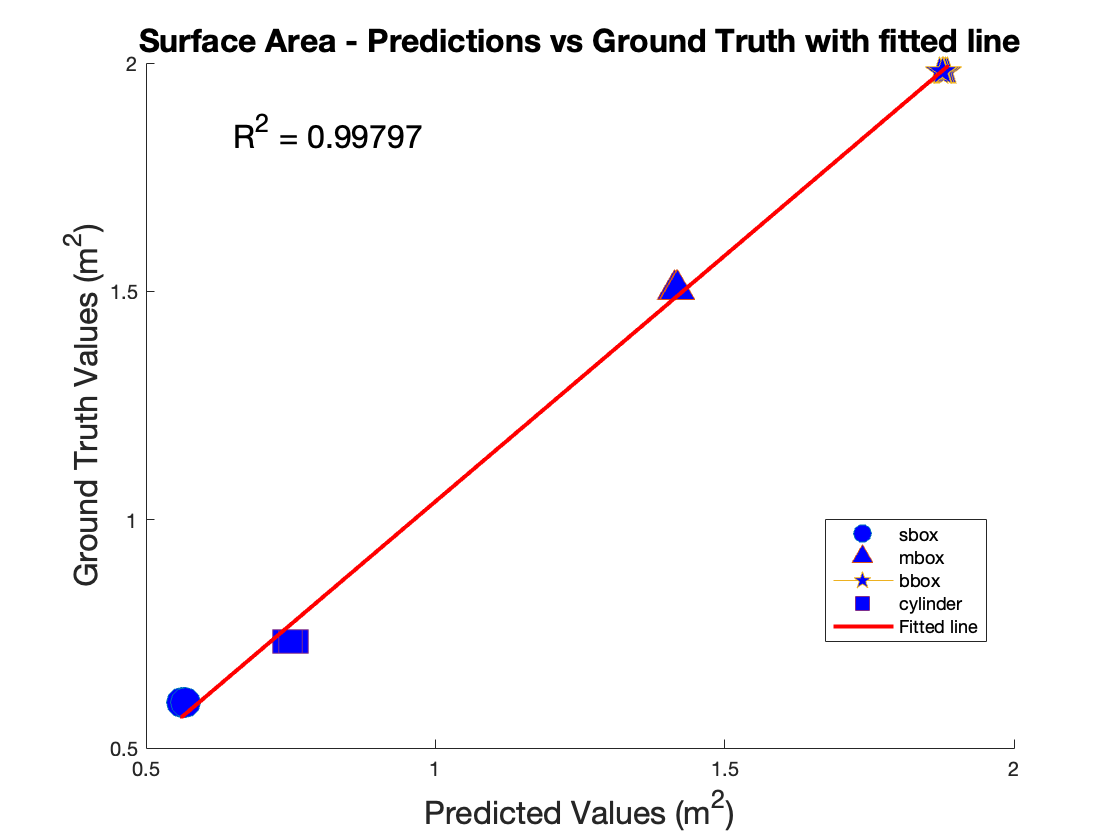}
         \caption{Surface area calculation results}
         \label{fig:mulsurres}
     \end{subfigure}
     \hfill
     \begin{subfigure}[b]{0.4\textwidth}
         \centering
         \includegraphics[width=\textwidth]{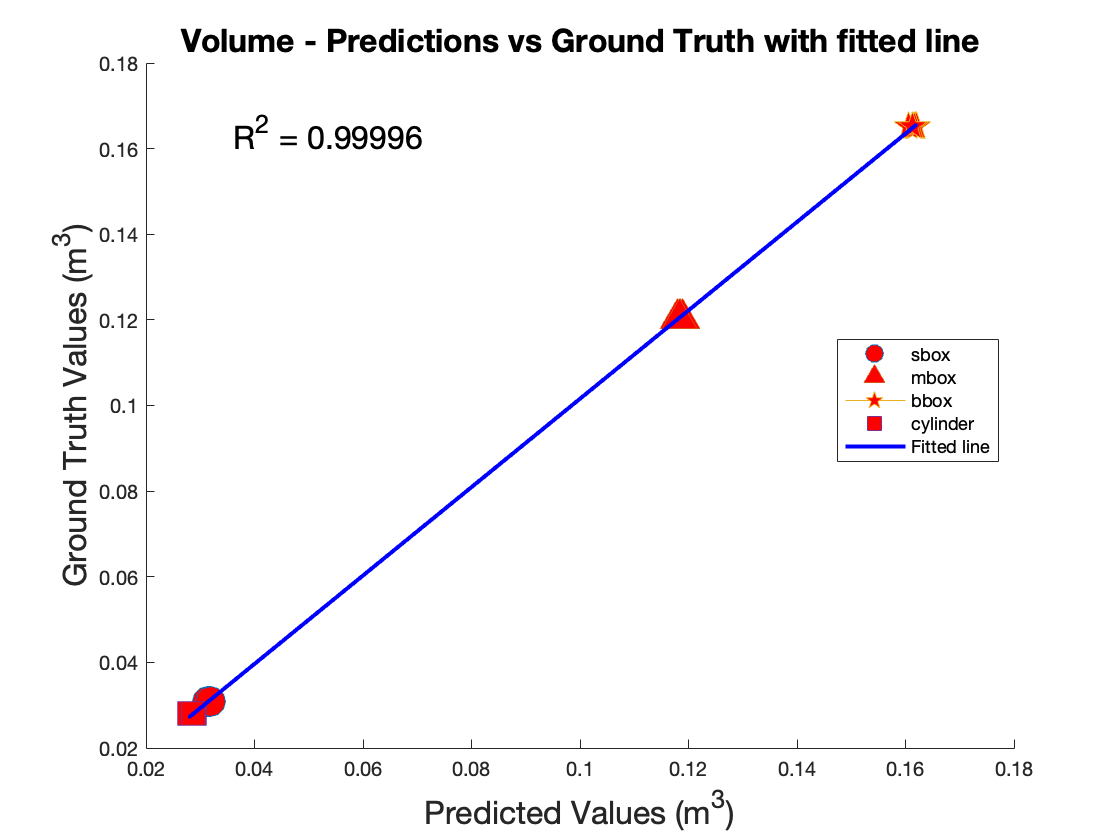}
         \caption{Volume calculation results}
         \label{fig:mulvolres}
     \end{subfigure}
     \hfill
     \caption{The statistical results of scanning different known objects.}
     \label{fig:mulres}
\end{figure}

The experimental investigation involving the cylindrical object, conducted across various orientations, established that the scanner's performance remains unaffected by changes in object pose. In order to further validate the system, the remaining objects were positioned in a consistent orientation and subjected to multiple scans. This rigorous approach served to demonstrate the robustness and reliability of the scanner under varying conditions. Figure \ref{fig:mulres} shows the statistical results of scanning different known objects. 

\subsection{Experiments on Real Cattle}

After verifying the validity of the scanner on the known objects, experiments on the real cattle have been executed. To start the pipeline, the segmentation network had to be fine-tuned on the prepared cattle data set first. 

\subsubsection{Fine-Tuning Mask R-CNN}

\begin{figure*}
     \centering
     \begin{subfigure}[b]{0.4\textwidth}
         \centering
         \includegraphics[width=\textwidth]{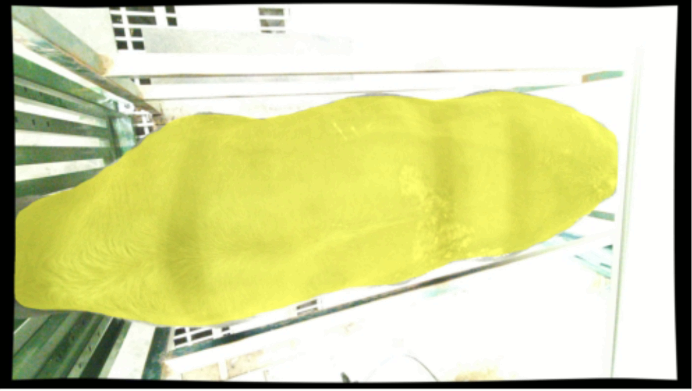}
         \caption{}
         \label{fig:segres1}
     \end{subfigure}
     \hfill
     \begin{subfigure}[b]{0.4\textwidth}
         \centering
         \includegraphics[width=\textwidth]{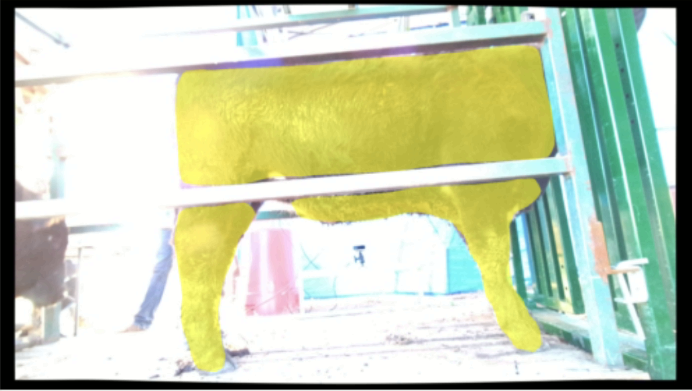}
         \caption{}
         \label{fig:segres2}
     \end{subfigure}
     \hfill
     \begin{subfigure}[b]{0.4\textwidth}
         \centering
         \includegraphics[width=\textwidth]{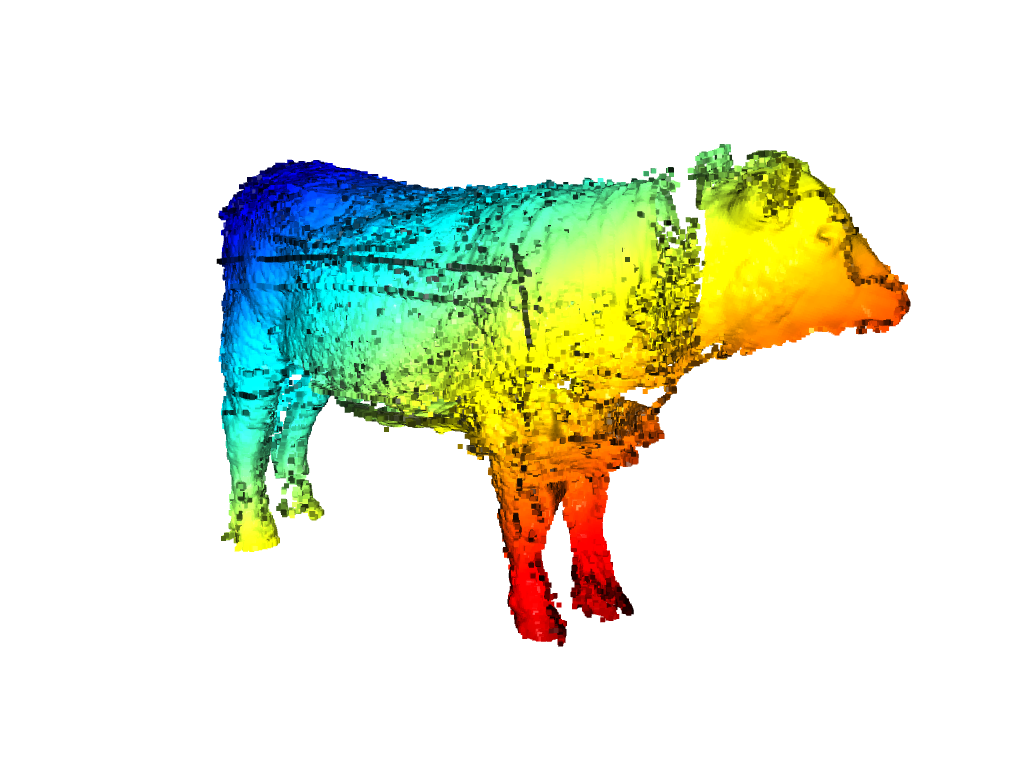}
         \caption{}
         \label{fig:regres3}
     \end{subfigure}
     \hfill
     \caption{Fine-tuning results of the Mask R-CNN on the cattle dataset prepared by the authors and a sample of registered animals.}
     \label{fig:segres}
\end{figure*}

As the camera sensors are providing depth and RGB images, two pre-trained Mask R-CNN was fine-tuned on depth and RGB images separately. Then the segmentation task was performed using voting arbitration between those two networks. Table \ref{tab:voting} shows the statistical results of using 4 different voting arbitration of segmentation models trained on the RGB and Depth images. As the results are revealing, using just the RGB model will produce the best average IOU score among the other arbitrations. However, it has a relatively high false negative rate which means there are multiple points on the cattle surface that are not being segmented as the animal correctly. This will cause the segmenting algorithm to lose some surface information that is useful to calculate the surface area of the animal. On the other hand, using single-vote arbitration is producing almost the same average IOU score, but it has the lowest false negative rate among the others. So the single-vote arbitration between the depth and RGB models has been chosen for the segmentation task.  Some qualitative segmentation results on the test set also have been depicted in Figure \ref{fig:segres}.

After successfully detecting and masking the animals within the RGB images, those results were used to clean and segment out the animal points from the point clouds. 

\begin{table}[]
    \centering
    \caption{The statistical results (the average IOU, the false positive rate (FR), and the false negative rate (FN)) of using different voting arbitration for the segmentation models. The bold font specifies the best scores among the others.}
    \begin{tabular}{|c|c|c|c|}
    \hline
    \textbf{Voting Arbitration} & \textbf{Avg. IOU} & \textbf{FP Rate (\%)} & \textbf{FN Rate (\%)} \\
    \hline
    RGB only & \textbf{0.9653} & 1.4480 & 3.7249 \\
    Depth only & 0.9495 & 2.6401 & 4.9985 \\
    1-Vote (OR) & \textbf{0.9629} & 3.2592 & \textbf{2.2631} \\
    2-Vote (AND) & 0.9518 & \textbf{0.7655} & 6.4603 \\
    \hline
    \end{tabular}
    \label{tab:voting}
\end{table}

\subsubsection{Registering All the Views}

In our study, the task of registering all the point clouds was done through a pairwise approach, utilizing the multi-scale Colored Iterative Closest Point (ICP) algorithm \cite{8237287}. The initial alignment of the point clouds in our method was facilitated through the use of a specially designed calibration cube. This cube was embedded with AprilTag fiduciaries, a type of 2D barcode, which provides reliable and accurate 3D pose estimation. This initial guess sets the stage for the more refined multi-scale colored ICP algorithm.

Subsequent to this initial alignment, the point clouds from each pair are registered using the colored ICP algorithm. The algorithm iteratively minimizes the difference between the points of two point clouds, simultaneously taking into account the color and geometric information. The beauty of this multi-scale approach is that it progressively refines the alignment starting from a coarse scale and gradually moving to a finer one, thereby ensuring a globally optimal solution while avoiding local minima.

Upon successful pairwise registration, all the registered pairs are linked together. This process effectively consolidates the registered point clouds from each individual camera perspective into one unified, globally consistent coordinate frame. The final, holistic point cloud brings together the different views, offering a comprehensive 3D representation of the object of interest. Figure \ref{fig:regres3} shows a registered point cloud of an animal after registering all the views. 


\subsection{Extracting Surface and Volume measurements on Cattle}

Experiments conducted on live cattle.

\begin{table*}[]
    \centering
    \begin{tabular}{|c|c|c|c|c|}
\hline
\textbf{ Cattle ID}	& \textbf{Manual-Measured Surface Area ($m^2$)}& 	 \textbf{Average Estimated Surface Area ($m^2$)} & \textbf{Surface Area Std ($m^2$)} & \textbf{Average Error (\%)}\\
\hline
1&  5.316&  5.52270& 0.09810 & 3.887\\
5&  5.088& 5.03896& 0.09579& 1.482\\ 
7&  5.189& 5.63482& 0.08180&  8.591\\
8&  5.202& 5.39062& 0.06005&  3.625\\
13& 5.660&  5.55332&  0.12395&  2.229\\
14& 5.406& 5.50881& 0.04657& 1.901\\
15& 5.024& 5.48532& 0.05472&  9.182\\
18& 5.329& 5.42638& 0.09839&  1.827\\
21& 5.278& 5.54710& 0.06522&  5.098\\
22& 5.571& 5.34380& 0.15375&  4.078\\
\hline
\textbf{Average}& - & - & \textbf{0.08783} & \textbf{4.1906}\\
\hline
    \end{tabular}
    \caption{Comparison between the measured surface area using the proposed 3D cattle scanner and the golden standard hand measurements. Each of the animals (which is determined by its cattle ID) was scanned 5 times and the average value was calculated over the scans.}
    \label{tab:my_label}
\end{table*}



\section{Conclusion and Future Works}

A new deep learning-based method for estimating cattle volume and surface area is introduced in this study. Of particular significance is the scanner's capability to extend beyond conventional domesticated livestock and encompass untamed animals, made possible by its utilization of an advanced deep-learning method for animal segmentation. By employing this cutting-edge approach, the proposed scanner emerges as an advantageous instrument for farmers and researchers in the realm of animal science, enabling rapid and accurate determination of volume and surface area for multiple animals within a remarkably short time frame. The resulting data holds immense potential for a wide range of applications, encompassing vital areas such as metabolism measurements and predictive modeling. Rigorous experimentation involving diverse objects not only affirmed the scanner's efficacy but also demonstrated its ability to deliver highly precise and reliable outcomes. Furthermore, extensive trials conducted on live animals provided compelling evidence of the scanner's remarkable reliability and accuracy, surpassing the conventional gold standard technique employed to measure the surface area of cattle. This groundbreaking system's distinctive ability to scan untamed animals, facilitated by its advanced deep learning-based segmentation, marks a pivotal advancement in animal scanning technology, opening new horizons for wildlife studies, ecological research, and population monitoring.

In light of the conclusion drawn from the aforementioned findings, several potential avenues for future research and development can be explored. Firstly, it would be beneficial to investigate the applicability of the proposed cattle volume and surface area scanner to different animal species, such as goats, sheep, or pigs. This extension would contribute to the versatility and broader utility of the system across various agricultural domains.

Additionally, the refinement of the deep learning algorithms employed by the scanner could enhance its performance even further. Exploring advanced techniques like transfer learning or incorporating larger and more diverse datasets could potentially improve the accuracy and generalization capabilities of the model. Conducting comparative studies with alternative state-of-the-art scanning methodologies would also be valuable to assess the scanner's competitiveness and identify areas for further improvement.

Lastly, given the increasing demand for sustainable agriculture practices, future research endeavors could explore the integration of the scanner's data with environmental and resource management systems. This integration would enable optimized feed rationing, waste reduction strategies, and improved overall resource allocation for livestock production, promoting both economic and ecological sustainability.

\bibliographystyle{IEEEtran}
\bibliography{cow_paper.bib}

\end{document}